\documentclass[preprint, nopreprintline ,12pt]{elsarticle}

\usepackage{amssymb}

\usepackage{lineno}
\modulolinenumbers[0]

\usepackage{hyperref}
\hypersetup{
    colorlinks=true
}

\usepackage[table, dvipsnames]{xcolor}
\usepackage{tikz}

\usepackage{geometry}
\usepackage{ulem} 
\usepackage{amsmath}
\usepackage{amsthm}
\usepackage{bbold}
\usepackage{bm}
\usepackage{booktabs}

\definecolor{cyan}{rgb}{0,0.5,.2}
\definecolor{dcyan}{rgb}{0,0.3,.3}

 \definecolor{pink}{HTML}{ff00d4}
 \definecolor{dpink}{HTML}{99007f}

\definecolor{lblue}{rgb}{0.9,0,0.5}
\definecolor{dblue}{rgb}{0.9,0,0.3}

\definecolor{maxd}{HTML}{D22E2E}
\definecolor{mind}{HTML}{F47B00}

\begin{document}

\journal{Energy and AI}

\begin{frontmatter}



	\title{High-Resolution Peak Demand Estimation Using Generalized Additive Models and Deep Neural Networks}

	\author[a]{Jonathan Berrisch}
	\ead{jonathan.berrisch@uni-due.de}

	\author[a]{Micha\l{} Narajewski}
	\author[a]{Florian Ziel}

	\address[a]{University of Duisburg-Essen, Germany}

	\begin{abstract}
		This paper covers predicting high-resolution electricity peak demand features given lower-resolution data.
		This is a relevant setup as it answers whether limited higher-resolution monitoring helps to estimate future high-resolution peak loads when the high-resolution data is no longer available. That question is particularly interesting for network operators considering replacing high-resolution monitoring predictive models due to economic considerations.
		We propose models to predict half-hourly minima and maxima of high-resolution (every minute) electricity load data while model inputs are of a lower resolution (30 minutes). We combine predictions of generalized additive models (GAM) and deep artificial neural networks (DNN), which are popular in load forecasting.
		We extensively analyze the prediction models, including the input parameters' importance, focusing on load, weather, and seasonal effects. The proposed method won a data competition organized by Western Power Distribution, a British distribution network operator. In addition, we provide a rigorous evaluation study that goes beyond the competition frame to analyze the models' robustness. The results show that the proposed methods are superior to the competition benchmark concerning the out-of-sample root mean squared error (RMSE). This holds regarding the competition month and the supplementary evaluation study, which covers an additional eleven months. Overall, our proposed model combination reduces the out-of-sample RMSE by 57.4\% compared to the benchmark.
	\end{abstract}




\end{frontmatter}

\newpage



\section{Introduction}
The decentralization of the energy system renders accurate electricity demand forecasts more critical than ever. Spikes in demand can produce strain on the networks. These issues will likely increase due to the increased use of lower-carbon technologies such as heat pumps and electric vehicles. Monitoring is expensive in the long term since it requires initial investments and sustained maintenance. Network operators are, therefore, particularly interested in peak-demand estimates.

This paper presents a winning method for high-resolution peak-demand prediction using generalized additive models (GAM) and deep neural networks (DNN). We developed this method for solving a data competition organized by Western Power Distribution (WPD), a British network operator responsible for the Midlands, South West, and Wales. The goal was to predict high-resolution minimum and maximum peak load using only low-resolution data and weather information. This is a relevant setup as it answers whether limited higher-resolution monitoring helps to estimate future high-resolution peak loads when the high-resolution data is no longer available. This is particularly interesting for network operators since consistent high-resolution monitoring is very costly \cite{zheng2013smart, suanduleac2021high}. {Therefore, replacing high-resolution monitoring with data science methods is an economically attractive scenario. If the results for a particular substation generalize to other substations, only low-resolution data would be necessary to obtain high-resolution features.}


Peak load estimates are relevant at various scales. While Nationwide forecasts are essential for climate change issues and resource planning~\cite{mughees2021deep, lee2022national}, network operators are usually more interested in regional peak load predictions, which are also the topic of this paper. Another relevant topic for network operators is peak load shaving~\cite{uddin2018review, lissa2021deep}. The latter deals with flattening the load curve. While this topic is also highly relevant for network operators, it is less related to the topic of this paper. In the following paragraphs, we provide an overview of research on peak load estimation.

There are various papers dealing with peak load prediction at different scales, e.g, forecasting short-term load of individual households~\cite{sun2018probabilistic, chou2018forecasting}, office buildings~\cite{chen2017short}, low-voltage feeders~\cite{haben2019short}, cities~\cite{guo2018deep}, and nations~\cite{lee2022national}. The results indicate that weather data like temperature, humidity, wind speeds, and air-pressure help explain peak loads~\cite{xie2016relative, dehalwar2016electricity, cai2019day, muzaffar2019short}. {The influence of windspeed on electricity loads is discussed in detail in \cite{acarouglu2021comprehensive}. \cite{hong2015weather} deals with the selection of suitable weather stations.} In contrast,~\cite{haben2019short}, who supplemented their model with deterministic factors like the day of the week, found little to no evidence for an impact of the temperature on electricity load.

The applied techniques for peak load forecasting are diverse. They include models like time-varying autoregressive (ARTV)~\cite{vu2017short}, recurrent and convolutional neural networks (RNN, CNN)~\cite{cai2019day}, long-short term memory (LSTM) networks~\cite{muzaffar2019short}, gradient boosting \cite{aguilar2021short}, and multistep approaches ~\cite{zhang2017short, sheng2020short, fan2021forecasting}. The techniques proposed in this paper have also been applied to load forecasting. GAMs are widely used in global~\cite{pierrot2011short} and local~\cite{goude2013local, ziel2022smoothed} short-term electricity load forecasting, and in forecasting high-resolution demand smart-meter data~\cite{amato2021forecasting}. The literature on artificial intelligence models based on neural networks in electricity load forecasting is much broader. A wide range of artificial neural networks, including multilayer perceptron networks (MLP), CNNs, and RNNs, are used in peak load~\cite{tasre2011daily, pallonetto2022forecast} and load~\cite{hosein2017load, amarasinghe2017deep, khwaja2020joint, walser2021typical, memarzadeh2021short, shaqour2022electrical, bashir2022short, khan2022efficient} forecasting.

Summarizing the research contribution of this manuscript, we
\begin{itemize}
	\item[i)]	propose a winning method for predicting high-resolution power load with low-resolution data.
	\item[ii)] analyze the descriptive data characteristics of relevant input data (esp. load and weather inputs) in detail for suitable usage for predictive analytics.
	\item[iii)] design machine learning models, GAMs and DNNs, which combined yield the best performance among competitors.
	\item[iv)] provide a rigorous predicting study that ensures the robustness of the method.
	\item[v)] study and interpret the explanatory power of relevant inputs and discuss hyperparameter tuning results.
\end{itemize}

The remainder of this paper is structured as follows. Section~\ref{data} introduces the data and provides an extensive descriptive data analysis. Section~\ref{models} describes the models that we developed. The models' performance is measured using an evaluation study. Section~\ref{results} contains the study design as well as the results. Section~\ref{conclusion} concludes.

\begin{figure}[bht!]
	\resizebox{\textwidth}{!}{
		\input{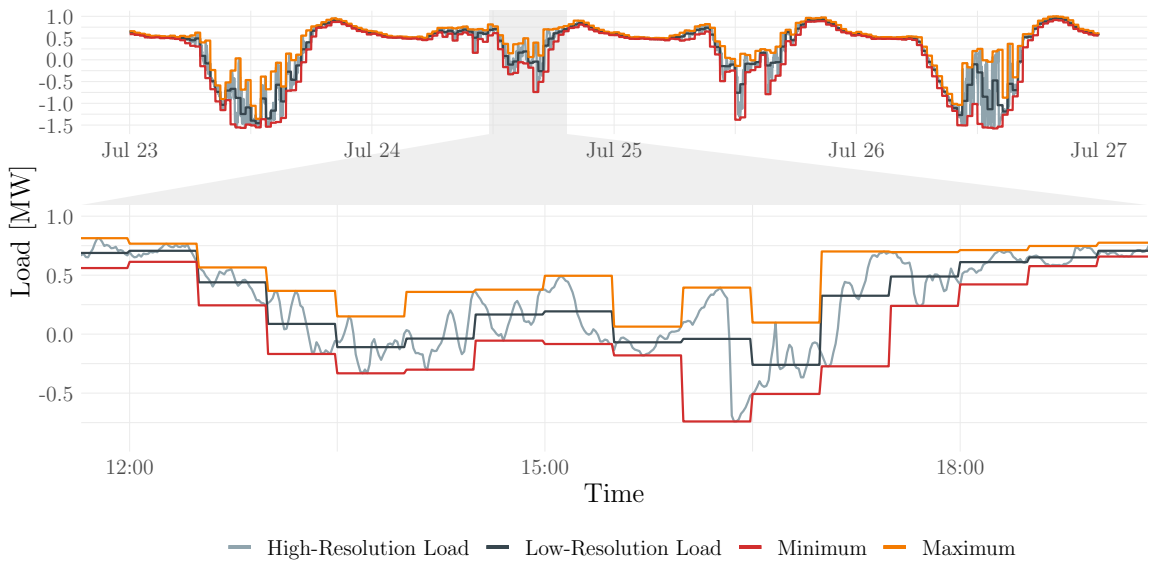}
	}
	\caption{Illustration of the high-resolution peak load estimation task for four days in July 2021. The half-hourly minima and maxima (red, orange) of the high-resolution 1-minute load data (grey) have to be predicted using the low-resolution half-hourly load values (black).}\label{fig_intro}
\end{figure}

\section{Setting, Data \& Exploratory Analysis}\label{data}
\begin{figure}
	\includegraphics[width=\textwidth]{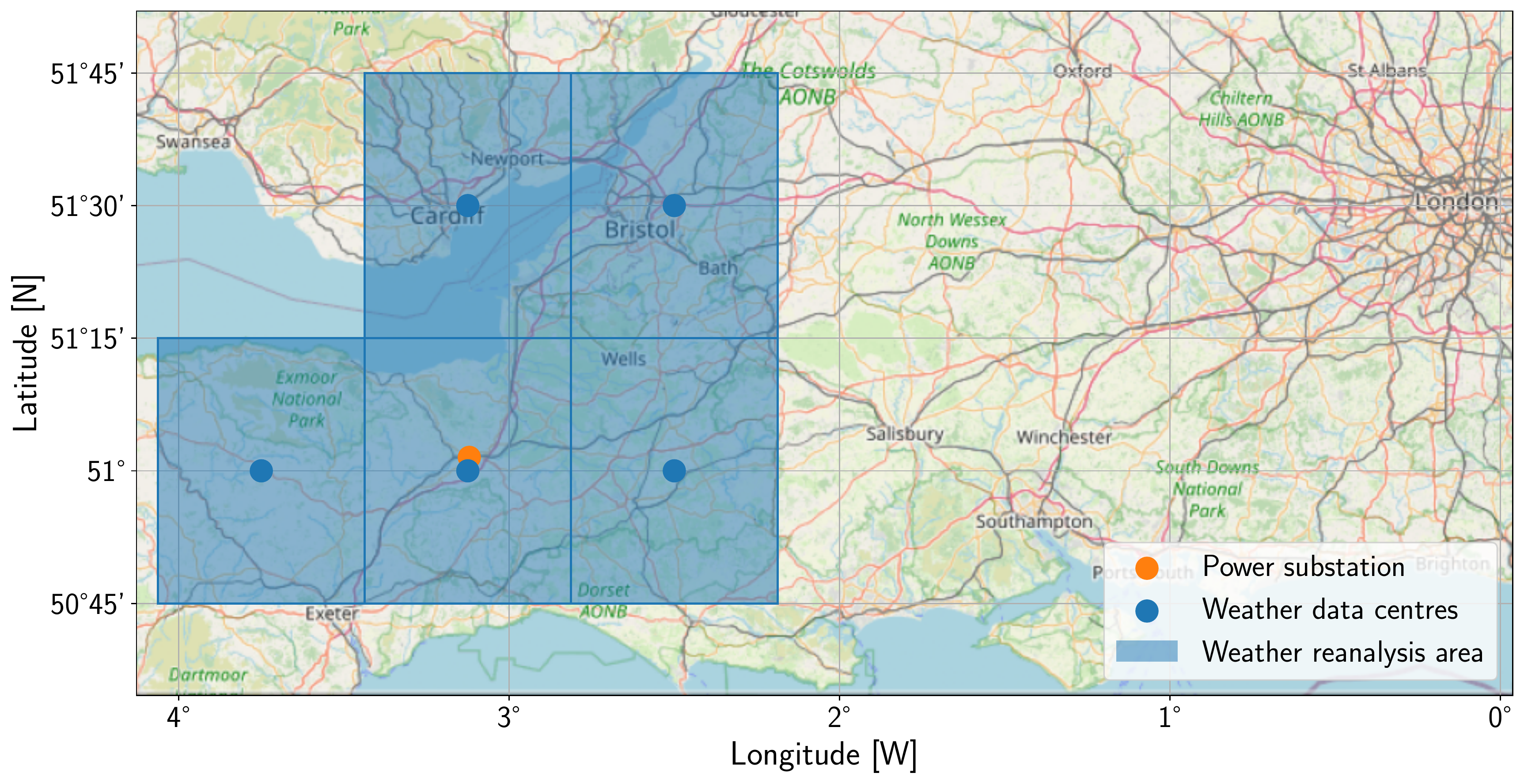}
	\caption{Location of the considered power substation and weather {reanalysis data} in the south-west of England. The map was generated using OpenStreetMap.}\label{fig_map}
\end{figure}

We consider a data set of the high-resolution peak load estimation challenge ranging from November 2019 to September 2021, covering almost two years of data. The data was provided by WPD\footnote{The data can be retrieved from \href{https://connecteddata.westernpower.co.uk/dataset/western-power-distribution-data-challenge-1-high-resolution-peak-estimation}{WPD Datasets}}.
Next to the mentioned load data that is also illustrated in Fig.~\ref{fig_intro}, weather data was given to improve predictions.
This is MERRA-2 weather reanalysis data from five {areas} close to the substation, as shown in Fig.~\ref{fig_map}. This weather data includes temperature (temp), solar irradiance (solar), north-south wind (windN), east-west wind (windE), pressure (press), and specific humidity (humid). We utilize the six weather inputs by averaging across the five {areas}.

Given the data mentioned above, the competition participants were requested to predict the minimum and maximum load values every half-hour of September 2021. However, to illustrate the robustness of the methodology used in the competition, we utilize a rolling window study with 12 test months from October 2020 to September 2021. Furthermore, the competition winner was determined by evaluating a specific skill with respect to the squared error. Thus, we are required to provide predictions of the expected value of the minimum and maximum load in each half-hour to optimize the skill score \cite{gneiting2011making}. Further details on the evaluation are described in Section~\ref{subsec_study}.

\begin{figure*}
	\resizebox{\textwidth}{!}{
		\input{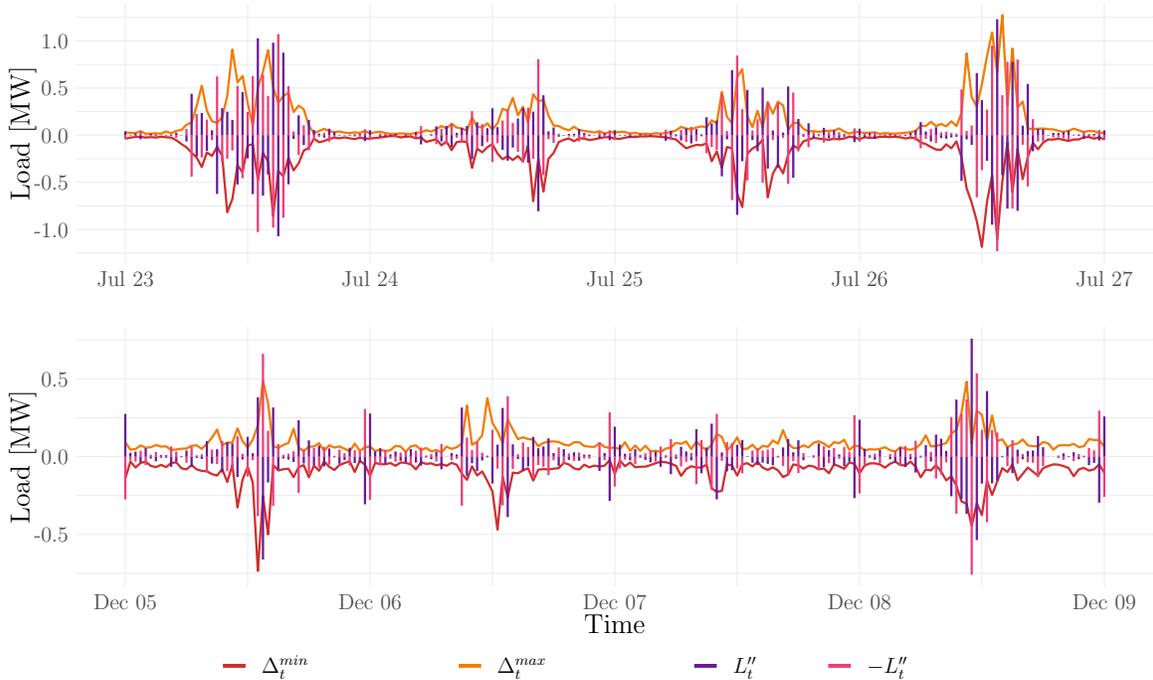}}
	\caption{Illustration of
		minimum and maximum load differences $\Delta^{\min}_t = L^{\min}_t - L_t$ and $\Delta^{\max}_t = L^{\max}_t - L_t$ with positive and negative discrete second order central difference (DSOCD) $L_t''$ of the load
		$L_t$ for four days in summer (23-26 July 2020, top) and winter (5-8 December 2020, bottom).}
	\label{fig_data1}
\end{figure*}

Let $L_t$ be the half-hourly load, corresponding to the black line in Fig.~\ref{fig_intro}. Further, let $L^{\min}_t$ and $L^{\max}_t$ be the minimum and maximum load values of the high-frequency data, which are our prediction targets. As $L_t$ is known and it holds $L^{\min}_t<L_t<L^{\max}_t$ it makes sense to model $\Delta^{\min}_t = L^{\min}_t - L_t $ and $\Delta^{\max}_t = L^{\max}_t - L_t $ directly. Fig.~\ref{fig_data1} illustrates the minimum and maximum load difference $\Delta^{\min}_t$ and $\Delta^{\max}_t$ for two selected time regions. Here, we added the discrete second order central difference (DSOCD) additionally
\begin{equation}
	L_t'' = L_{t-1}-2L_t+L_{t+1}
\end{equation}
of the load $L_t$, as well as its negative version $-L_{t}''$. We observe that $\Delta^{\min}_t$ and $\Delta^{\max}_t$ tends to be larger in absolute terms if $L_t''$ (or a neighbouring value like $L_{t-1}''$ or $L_{t+1}''$) is large in absolute terms. Thus, it makes sense to regard $L_t''$ as an important feature in model engineering.

A closer look at the bottom graph in Fig.~\ref{fig_data1} shows relatively large absolute values for $L_t''$ in winter at the very end and beginning of a day (at 23:30 and 0:00). {It concerns a limited period in the winter months exclusively. However, we can not explain those values.} In contrast to other large values, those do not translate into large absolute $\Delta^{\min}_t$ and $\Delta^{\max}_t$ values. {Therefore, we decided to replace these values to prevent any influence of these spurious observations on our prediction.} To adjust for these phenomena, we introduce an adjusted discrete second-order central difference $\widetilde{L}_t''$ where the corresponding linear interpolation replaces the two values. The latter is formally defined as:
\begin{align}
	\widetilde{L}_{t}'' = \begin{cases}
		                      L_{t}''                                       & \text{if } t \text{ mod } 48 = 1, \ldots ,46 \\
		                      \frac{1}{3} L_{t-2}'' + \frac{2}{3} L_{t+1}'' & \text{if } t \text{ mod } 48 = 0             \\
		                      \frac{2}{3} L_{t-2}'' + \frac{1}{3} L_{t+1}'' & \text{if } t \text{ mod } 48 = 47
	                      \end{cases} .
\end{align}

We define three deterministic components for further preliminary data analysis to capture potential seasonal characteristics in the data. Those are daily, weekly and annual inputs, which we refer to by $D_t$, $W_t$, and $A_t$. $D_t$ and $W_t$ count the number of hours in a day or week starting  Monday. Similarly, $A_t$ counts the number of hours in a meteorological year with $365.24$ days starting at 0:00 on 1st January 2020.

\begin{figure}
	\resizebox{\textwidth}{!}{
		\input{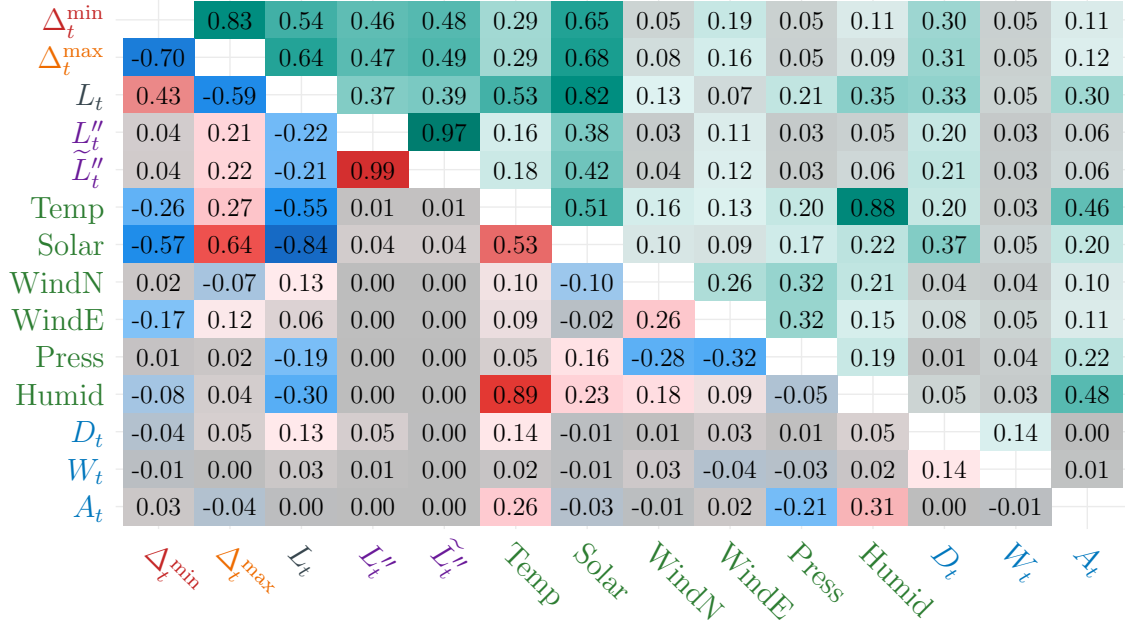}
	}
	\caption{Correlation plot with Pearson's correlation on the lower triangle and distance correlation on the upper triangle.}\label{fig_data2}
\end{figure}

Fig.~\ref{fig_data2} shows a correlation matrix of relevant data discussed above. These are load-related time series $\Delta^{\min}_t$, $\Delta^{\max}_t$, $L_t$, $\widetilde{L}_t''$ $L_t''$, the six weather time series, and the daily, weekly and annual seasonal inputs $D_t$, $W_t$, and $A_t$. We illustrate Pearson's correlation coefficients on the lower triangle and the distance correlation on the upper triangle. The former is a standard linear dependency measure that takes values in $[-1,1]$. The latter is a non-linear dependency measure that takes values in $[0,1]$ and characterizes stochastic independence~\cite{szekely2007measuring}.

We observe that likely $L_t$ and $\widetilde{L}_t''$ will be important to explain $\Delta^{\min}_t$ and $\Delta^{\max}_t$. We also see that $\widetilde{L}_t''$ seems to be slightly preferable to $L_t''$ for modeling purposes. In terms of correlation, solar irradiance explains best $\Delta^{\min}_t$ and $\Delta^{\max}_t$ among all meteorological inputs, which is plausible due to the expected high impact from photovoltaic to the load. However, the overall contribution of the weather data remains unclear, especially as weather data is highly correlated with deterministic seasonal variables. Among the seasonal inputs, we observe close-to-zero correlation values between $-0.04$ and $0.05$. This means that there is merely and linear dependency between $\Delta^{\min}_t$ and $\Delta^{\max}_t$ and the seasonal inputs. However, we observe elevated distance correlations for the daily and annual inputs. Since strong linear relations are already ruled out due to the close-to-zero Pearson's correlation, we can deduce that the relation between daily and annual inputs and $\Delta^{\min}_t$ and $\Delta^{\max}_t$ is non-linear.

\section{Models}\label{models}

To preserve comparability, we design the GAM and  DNN models using the same input matrix $\mathbb X$.
$\mathbb X$ contains $N=21$ columns that are listed in Tab.~\ref{tab_inputs}. Note again that the challenge did not ask for forecasts (of the future) but for precise predictions. Therefore, in contrast to a classical forecasting problem, it is allowed to use positively lagged inputs to deduce the high-frequency peak load behavior. Therefore, we include positive lags of the load and DSOCD load.\footnote{However, at the end of the data set, we have to omit positively lagged components to predict the last observations of September 2021.}
\begin{table}
	\begin{tabular}{llc}
		\toprule
		Variable type     & Included feature                                       & Number \\ \midrule
		Lagged load       & $L_{t-1}$, $L_t$, $L_{t+1}$                            & 3      \\
		Lagged DSOCD load & $\widetilde{L}_{t-4}'',\ldots , \widetilde{L}_{t+4}''$ & 9      \\
		Weather inputs    & Temp, Solar, WindN, WindE, Press, Humid                & 6      \\
		Seasonal inputs   & $D_t$, $W_t$, $A_t$                                    & 3      \\    \bottomrule
	\end{tabular}
	\caption{Table of all 3+9+6+3=21 inputs.}\label{tab_inputs}
\end{table}

\subsection{GAM models}

A generalized additive model (GAM) is a model with $M$ additive model components.
For our objectives $\Delta_t^{\min}$ and $\Delta_t^{\max}$ this is in general
\begin{equation}
	\Delta^{m}_t = \sum_{i=1}^M f_i( X_{t,1}, \ldots, X_{t,N}) +\varepsilon_t
	\label{eq_GAM_gen}
\end{equation}
where $m\in\{\min , \max\}$. The additive terms $f_i$ may represent arbitrary model components.
Traditionally, linear terms and smoothing spline components are used for $f_i$. However, complex building blocks like gradient boosting machines or (deep) neural networks are also used in recent years, \cite{ziel2021m5, marz2022distributional}.

In this application, we reduce the complexity of the considered GAM model~\eqref{eq_GAM_gen} by restricting the model to linear terms and 2-way interactions. Thus, we do not allow for 3-way or higher interactions between the inputs. Additionally, we use GAMs in the traditional framework using splines, more precisely cubic B-splines. The GAM model with all 2-way interactions is defined by
\begin{equation}
	\Delta^{m}_t = \sum_{i=1}^N b_{k_0}( X_{t,i} ) + \sum_{i=1}^N \sum_{j=1,j>i}^N b_{k_1,k_2}( X_{t,i}, X_{t,j} ) +\varepsilon_t
	\label{eq_big_GAM}
\end{equation}
where $b_{k_0}$ and $b_{k_1,k_2}$ denote univariate and bivariate splines with $k_0$, and $(k_1,k_2)$ knots. For the latter, we consider tensor interaction splines that only model joint interaction effects.
In addition, we choose $k_0=27$ and $k_1=k_2=9$. Thus, linear terms are specified by $27$ parameters and bivariate terms by $81$ parameters.

The model is estimated using \texttt{bam} function of the \texttt{R} package \texttt{mgcv}~\cite{wood2017gam} which allows efficient estimation for large data sets. Here, the input data can be discretized first to reduce the computational effort in the fast restricted maximum likelihood (fREML) computation. The fREML estimation is performed with respect to the Gaussian distribution with fixed variance, which corresponds to minimizing least squares.
The parameter estimation is performed so that the Bayesian information criterion (BIC) is minimized, utilizing the effective degrees of freedom of the smoothing spline components.
For more details on computational and theoretical issues on spline-based GAM models see~\cite{wood2017gam}.

In addition to~\eqref{eq_big_GAM}, we introduce a reduced GAM model (\textbf{GAM.red}) that includes only selected 2-way interactions. We consider here only interactions that involve the crucial
load input $L_t$, the second derivative $\widetilde{L}_t''$ and the daily seasonal term $\text{Solar}_t$.
Thus, the reduced GAM model is
\begin{align}
	\Delta^{m}_t & = \sum_{i=1}^N b_{k_0}( X_{t,i} )
	+ \sum_{i=2}^N b_{k_1,k_2}( L_t, X_{t,i} ) \nonumber                    \\
	             & + \sum_{i=3}^N b_{k_1,k_2}( \widetilde{L}_t'', X_{t,i} )
	+ \sum_{i=4}^N b_{k_1,k_2}( \text{Solar}_t, X_{t,i} )
	+\varepsilon_t
	\label{eq_small_GAM}
\end{align}
with the same $k_0=27$ and $k_1=k_2=9$. Note that for notation convenience in~\eqref{eq_small_GAM} we assume that in column 1 to 3 in $\mathbb X$ are $L_t$, $\widetilde{L}_t''$ and $\text{Solar}_t$.
The computational cost for parameter estimation of the \textbf{GAM.red} is substantially lower than in~\eqref{eq_big_GAM}. Using our eight-core machine, it relieves from about 3 hours to 2 minutes for the entire training and prediction of a minimum or maximum peak model.
	{We also considered the main GAM models \textbf{GAM.full} and \textbf{GAM.red} with absolute wind speed\footnote{Absolute wind speed is defined as $\text{wind} = \sqrt{\text{windE}^2 + \text{windN}^2}$} instead of north-south and east-west components. However, the results differed only marginally (less than $0.05\%$ in RMSE). We, therefore, decided to use the original wind components and omit to report the results for brevity\footnote{We are happy to provide them upon request.}. The highly nonlinear structure of DNNs allows them to consider these nonlinear transformations themselves. We, therefore, decided to consider the original wind components only.}

\begin{figure}[htb!]
	\includegraphics[width=0.32\textwidth]{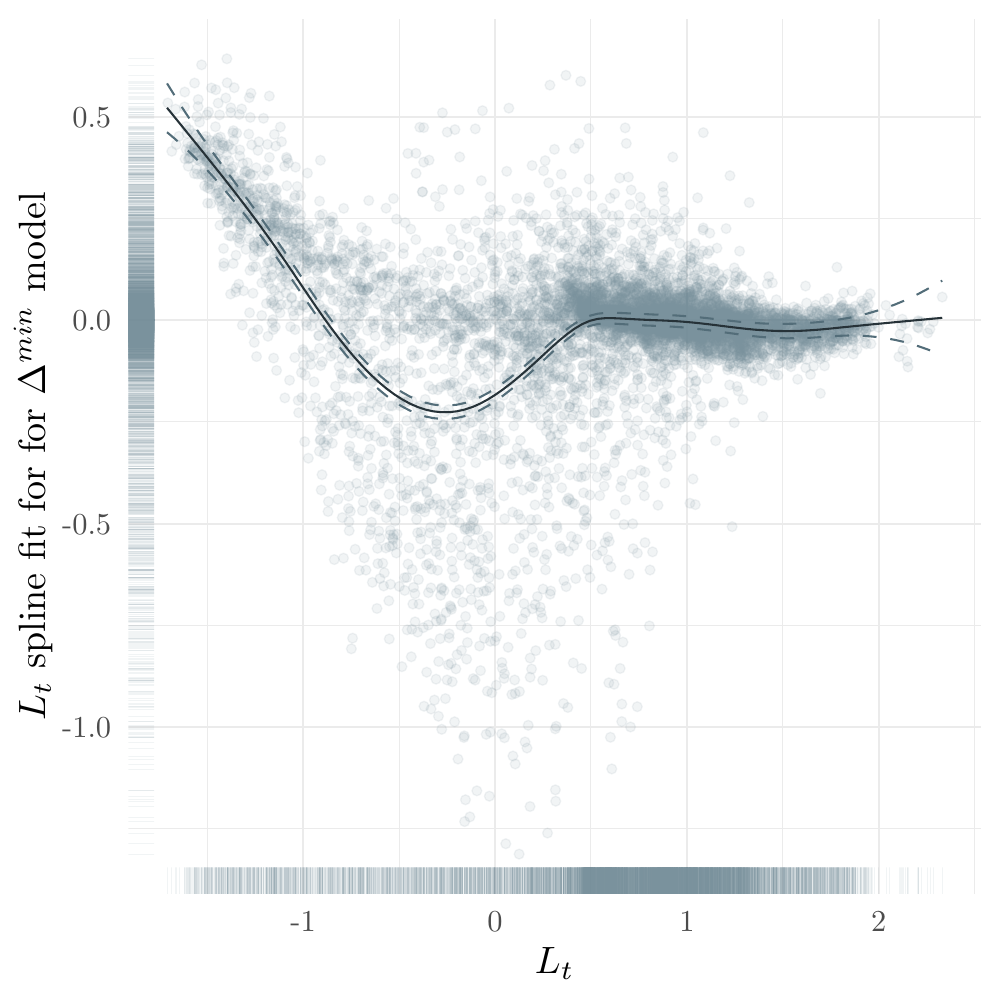}
	\includegraphics[width=0.32\textwidth]{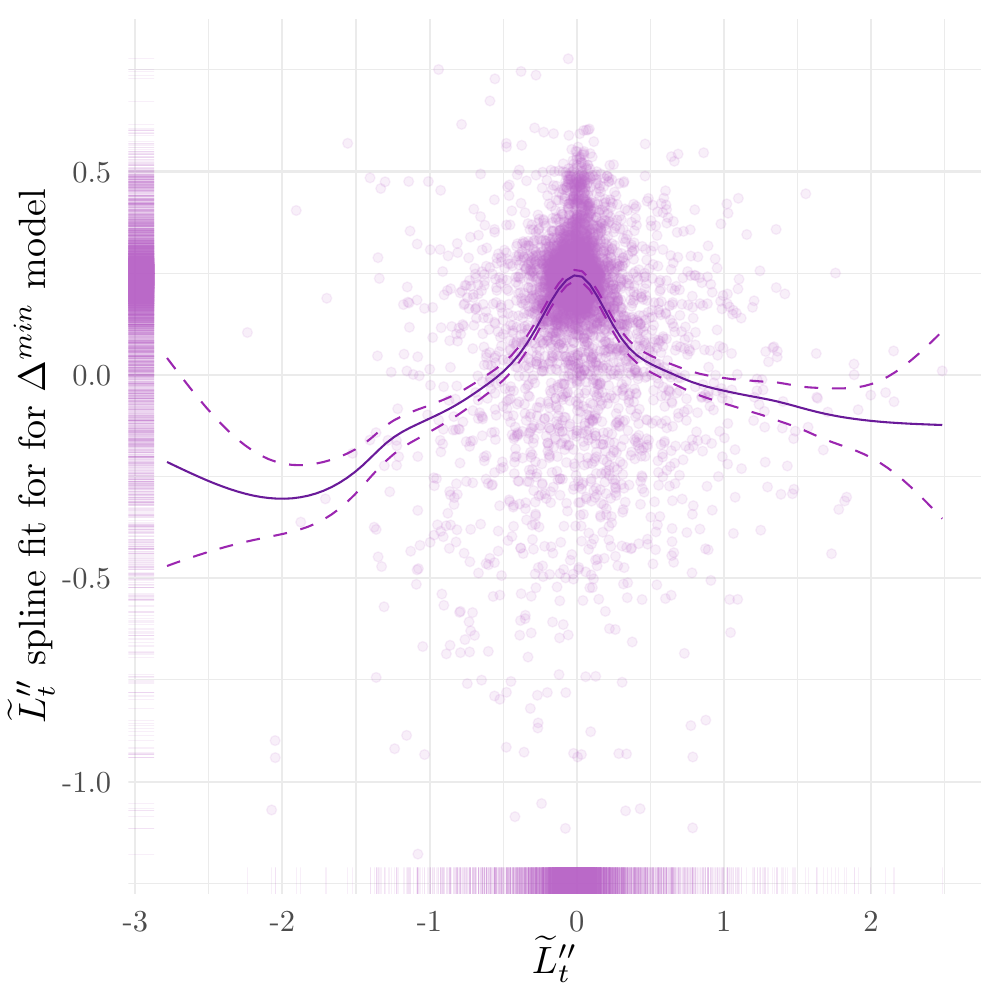}
	\includegraphics[width=0.32\textwidth]{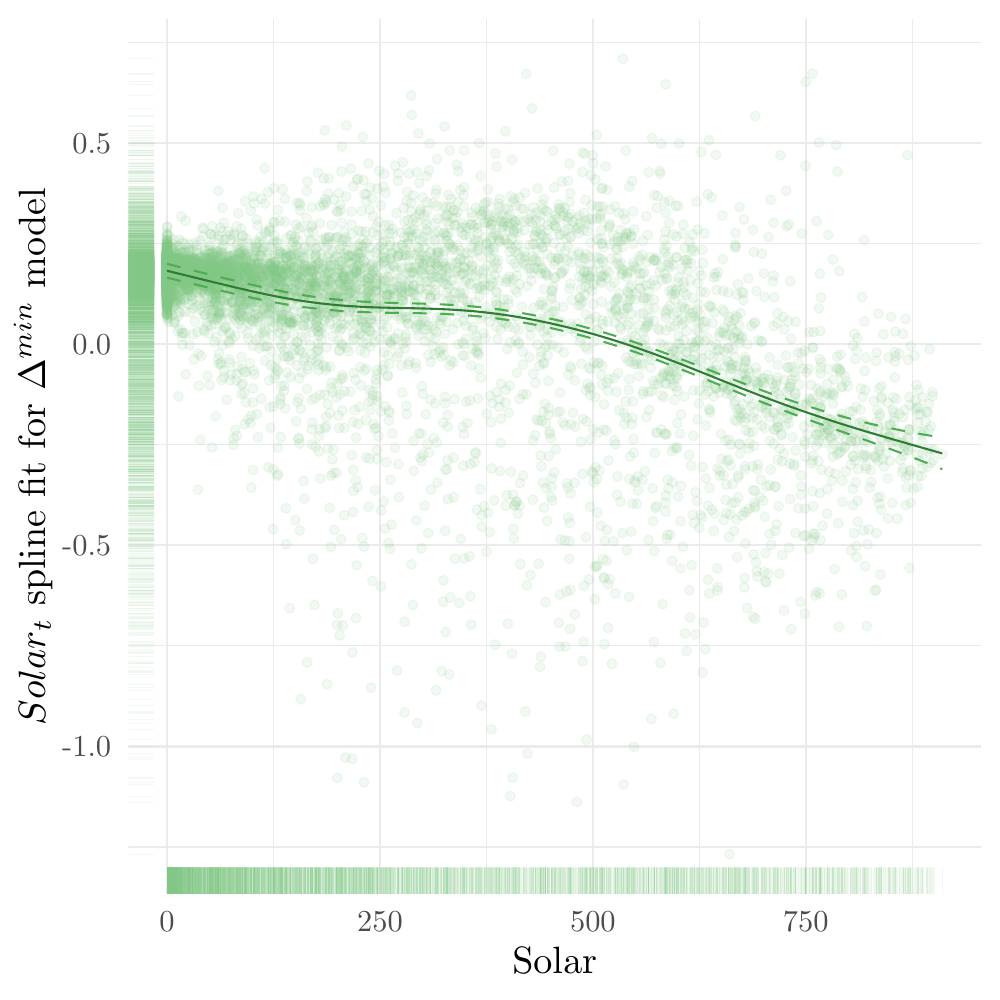} \\
	\includegraphics[width=0.32\textwidth]{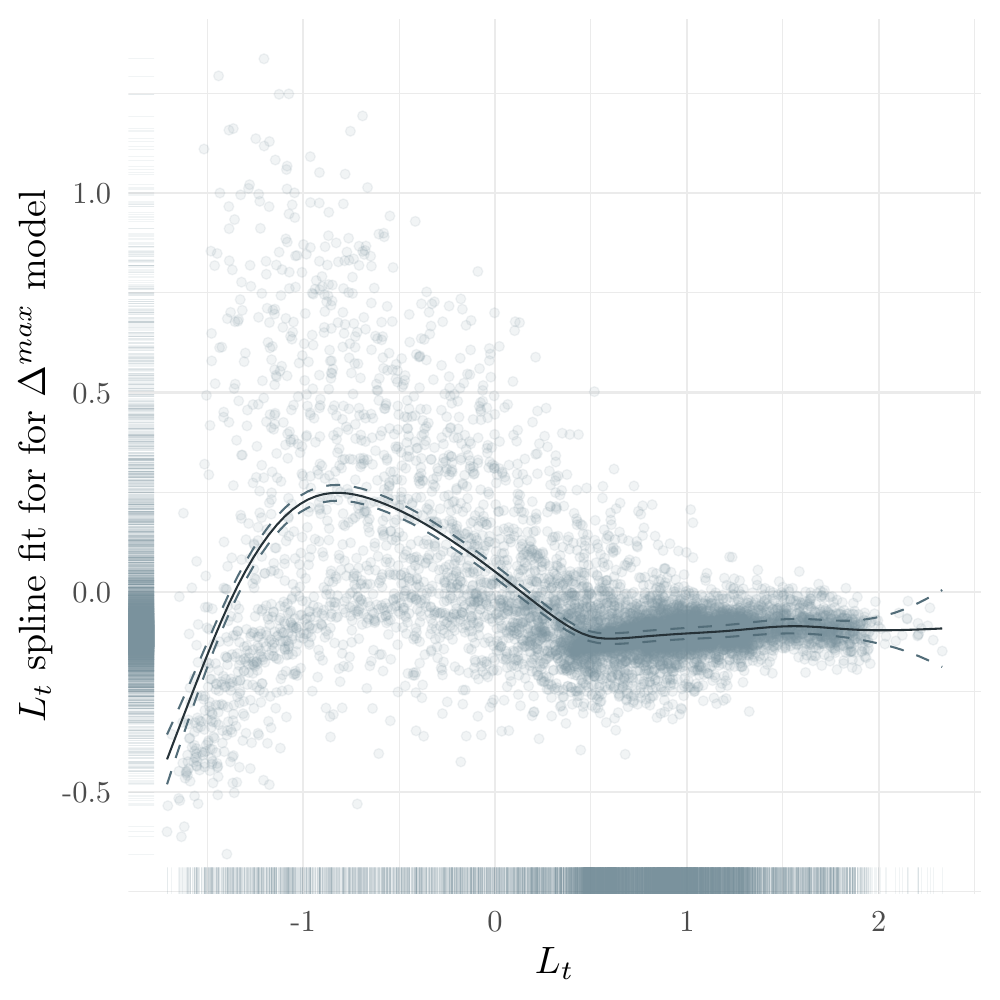}
	\includegraphics[width=0.32\textwidth]{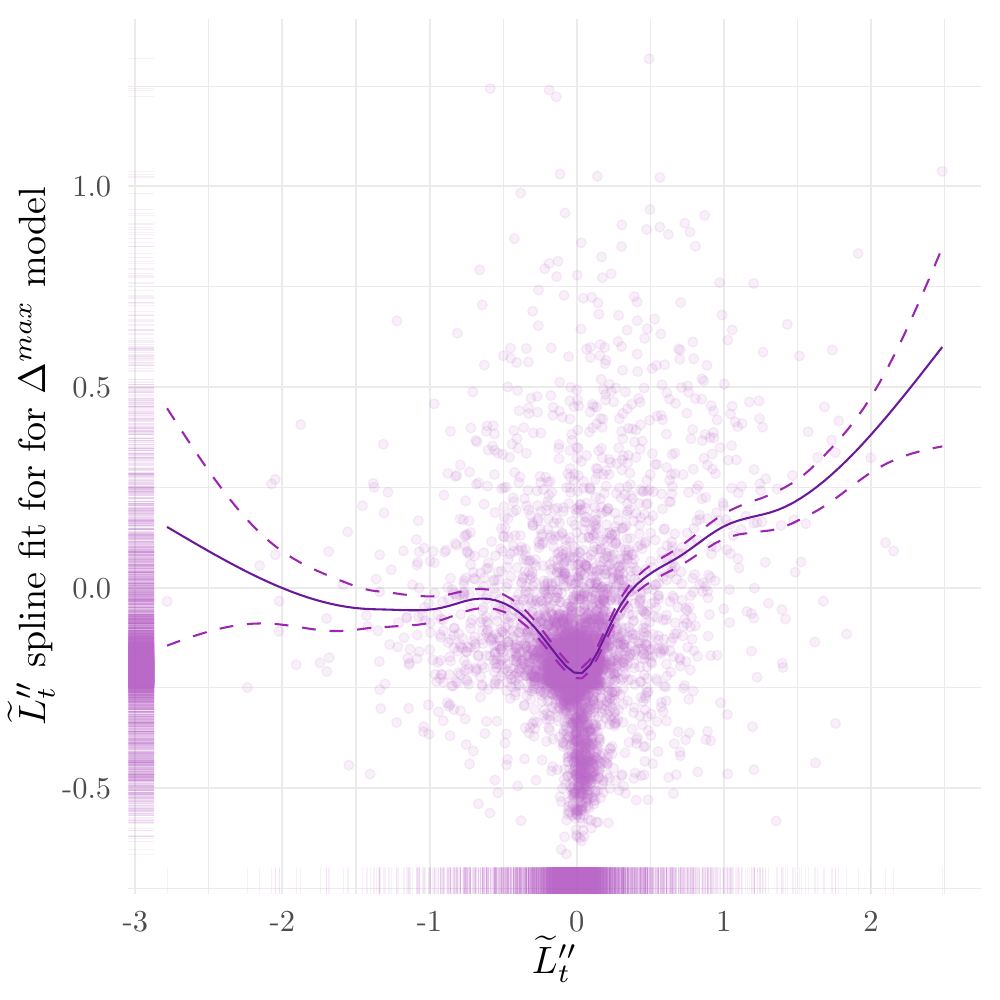}
	\includegraphics[width=0.32\textwidth]{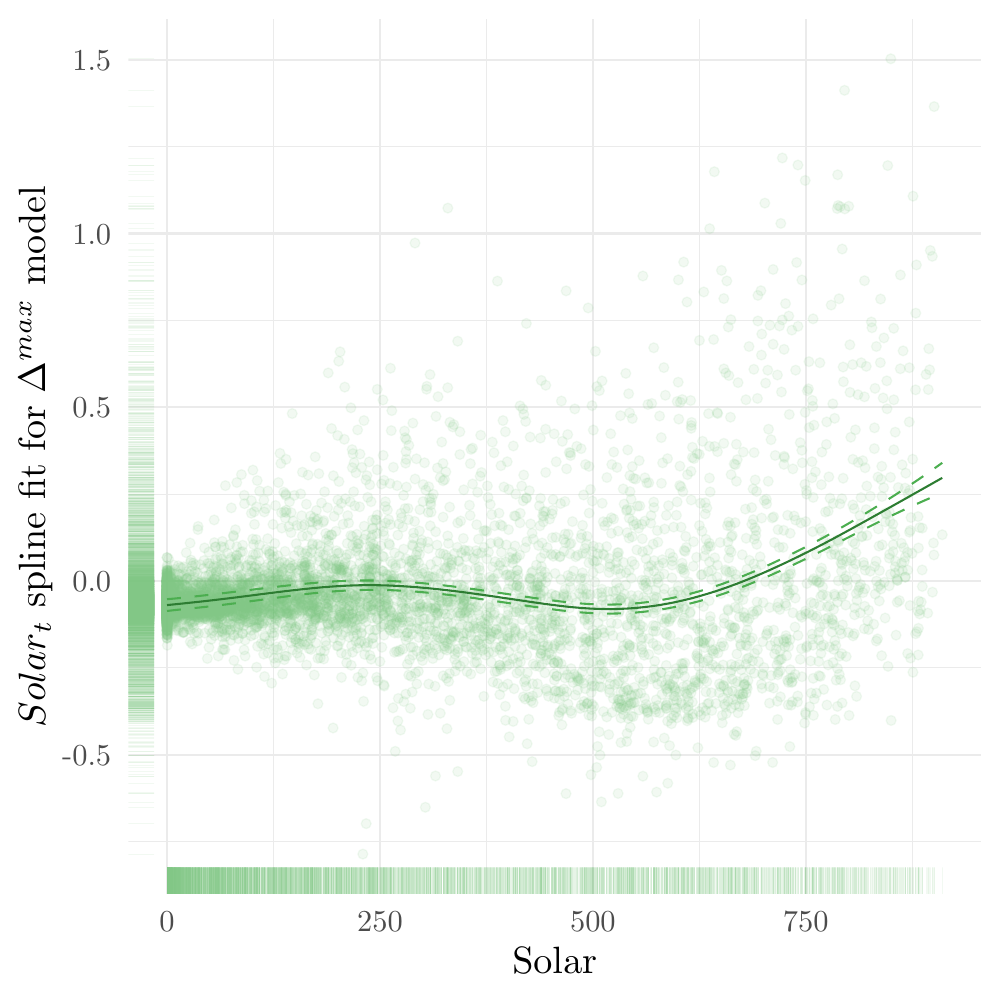}
	\caption{Fitted cubic B-splines of the simple GAM model
		\eqref{eq_simple_GAM} for $\Delta^{\min}_t$ (top) and $\Delta^{\max}_t$ (bottom) for $L_t$ (left), $\widetilde{L}_t''$ (center) and $\text{Solar}_t$ (right). }
	\label{fig_gam_graph}
\end{figure}

We consider the simple GAM model \textbf{GAM.simple} to illustrate the relevance and the non-linear impact of the mentioned three inputs $L_t$, $\widetilde{L}_t''$ and $\text{Solar}_t$ beyond the descriptive data analysis:
\begin{align}
	\Delta^{m}_t =  b_{k_0}( L_t ) +  b_{k_0}( \widetilde{L}_t'') + b_{k_0}( \text{Solar}_t )
	+\varepsilon_t
	\label{eq_simple_GAM}
\end{align}
with $k_0=27$. The BIC-based fit of model~\eqref{eq_simple_GAM} illustrated in Fig~\ref{fig_gam_graph}. We observe that especially the
load $L_t$ and the second derivative $\widetilde{L}_t''$ have clearly non-linear relationships. The spline fit of $\widetilde{L}_t''$ close to 0  looks approximately like an absolute value which corresponds to the effect that was already visible in Fig~\ref{fig_data1}.


Finally, we also consider the GAM model~\eqref{eq_big_GAM} without any weather inputs to evaluate the impact of the weather inputs. We refer to that model as \textbf{GAM.noWeather}.

\subsection{Deep neural network model}

\begin{figure}[h!]
	\tikzset{%
		input neuron/.style={
				circle,
				draw,
				minimum size=0.7cm
			},
		every neuron/.style={
				circle,
				draw,
				minimum size=.7cm
			},
		optional neuron/.style={
				circle,
				color=gray,
				draw,
				minimum size=.7cm
			},
		neuron missing/.style={
				draw=none,
				scale=1.2,
				text height=.25cm,
				execute at begin node=\color{black}$\vdots$
			},
		sigmoid/.style={path picture= {
						\begin{scope}[x=.7pt,y=7pt]
							\draw plot[domain=-6:6] (\x,{1/(1 + exp(-\x))-0.5});
						\end{scope}
					}
			},
		linear/.style={path picture= {
						\begin{scope}[x=5pt,y=5pt]
							\draw plot[domain=-1:1] (\x,\x);
						\end{scope}
					}
			},
		min/.style={path picture= {
						\begin{scope}[x=5pt,y=5pt]
							\draw node {\scriptsize{\textcolor{white}{$\Delta^{\text{max}}_t$}}};
						\end{scope}
					}
			},
		max/.style={path picture= {
						\begin{scope}[x=5pt,y=5pt]
							\draw node {\scriptsize{\textcolor{white}{$\Delta^{\text{min}}_t$}}};
						\end{scope}
					}
			},
	}

	\centering

	\begin{tikzpicture}[x=1cm, y=.65cm, >=stealth, scale=1.5]

		\node [input neuron/.try, neuron 1/.try] (input-1) at (-1,2.5-1) {\makebox[8.5pt]{}};
		\node [input neuron/.try, neuron 2/.try] (input-2) at (-1,2.5-2) {\makebox[8.5pt]{}};
		\node [input neuron/.try, neuron 3/.try] (input-3) at (-1,2.5-3) {\makebox[8.5pt]{}};
		\node [input neuron/.try, neuron 4/.try] (input-4) at (-1,2.5-4) {\makebox[8.5pt]{}};
		\node [input neuron/.try, neuron 5/.try] (input-5) at (-1,2.5-5) {\makebox[8.5pt]{}};
		\node [input neuron/.try, neuron 6/.try] (input-6) at (-1,2.5-6) {\makebox[8.5pt]{}};

		\foreach \m [count=\y] in {1,2,3,4}
		\node [every neuron/.try, neuron \m/.try, fill= black!20, sigmoid ] (hidden-\m) at (1,2.33-\y*1.33) {};
		\foreach \m [count=\y] in {1,2,3}
		\node [every neuron/.try, neuron \m/.try, fill= black!20, sigmoid ] (hidden2-\m) at (2.5,2-\y*1.5) {};

		\foreach \m [count=\y] in {1,2,3}
		\node [optional neuron/.try, neuron \m/.try, fill= gray!20, solid, sigmoid ] (hidden3-\m) at (4,2-\y*1.5) {};

		\foreach \m [count=\y] in {1}
		\node [every neuron/.try, neuron \m/.try, fill= maxd, max ] (output-1) at (5.5,1-\y) {};
		\foreach \m [count=\y] in {1}
		\node [every neuron/.try, neuron \m/.try, fill= mind, min ] (output-2) at (5.5,-1-\y) {};

		\foreach \l [count=\i] in {$X_{t,1}$,$X_{t,2}$,$X_{t,3}$,$X_{t,4}$,\ldots,$X_{t,N}$}
		\draw [<-] (input-\i) -- ++(-1.2,0)
		node [above, midway] {\footnotesize\l};

		\foreach \l [count=\i] in {1,2,3,4}
		\node [above] at (hidden-\i.north) {};


		\foreach \i in {1,...,6}
		\foreach \j in {1,...,4}
		\draw [->] (input-\i) -- (hidden-\j);

		\foreach \i in {1,...,4}
		\foreach \j in {1,...,3}
		\draw [->] (hidden-\i) -- (hidden2-\j);

		\foreach \i in {1,...,3}
		\foreach \j in {1,...,3}
		\draw [->] (hidden2-\i) -- (hidden3-\j);

		\foreach \i in {1,...,3}
		\foreach \j in {1,2}
		\draw [->] (hidden3-\i) -- (output-\j);

		\foreach \l [count=\x from 1] in {Hidden, Hidden}
		\node [align=center, above] at (\x*1.5-0.5,2) {\footnotesize\l \\[-3pt] \footnotesize{layer}};
		\foreach \l [count=\x from 3] in {Hidden}
		\node [align=center, above, color=gray] at (\x*1.5-0.5,2) {\footnotesize\l \\[-3pt] \footnotesize{layer}};
		\node [align=center, below, color=gray] at (3*1.5-0.5,2.25) {\footnotesize{(optional)}};
		\foreach \l [count=\x from 4] in {Output}
		\node [align=center, above] at (\x*1.5-0.5,2) {\footnotesize\l \\[-3pt] \footnotesize{layer}};
		\node [align=center, above] at (-1,2) {\footnotesize Input \\[-3pt] \footnotesize{layer}};

		\draw[dashed, color=gray] (3.4,2-3*1.5-1.5) -- (3.4, 3.2);
		\draw[dashed, color=gray] (4.6,2-3*1.5-1.5) -- (4.6, 3.2);
		\draw[dashed, color=gray] (4.6,2-3*1.5-1.5) -- (3.4, 2-3*1.5-1.5);
		\draw[dashed, color=gray] (4.6,3.2) -- (3.4, 3.2);

	\end{tikzpicture}
	\caption{Exemplary network structure of the MLP.}
	\label{fig_MLP}
\end{figure}
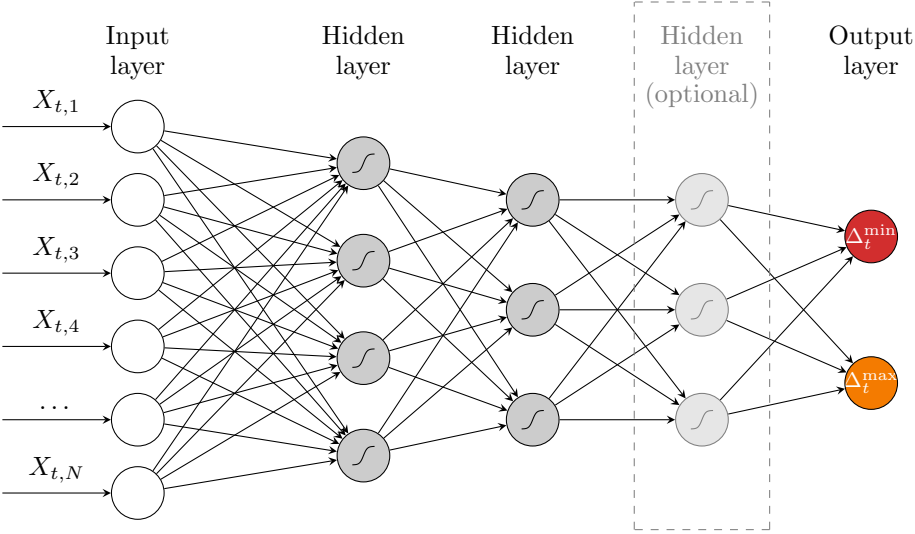

The deep neural network (DNN) model uses the MLP structure as shown in Fig.~\ref{fig_MLP}. Formally, for $i \in \{1, \dots, I\}$ we define it by
\begin{equation}
	\bm{H}_{t,i} = a_i\left(\bm{H}_{t,i-1}\bm{W}_{i} + \bm{b}_{i}\right)
	\label{eq_MLP}
\end{equation}
where $\bm{H}_{t,i}$ is the output matrix of $i$th hidden layer, and $\bm{W}_{i}$, $\bm{b}_{i}$, and $a_i$ are the weights, bias and activation function of the $i$th hidden layer, respectively. Note that $\bm{H}_{t,0} = \boldsymbol X_t = (X_{t,1},\ldots, X_{t,N})$ and $\bm{H}_{t,I+1} = \bm{\Delta}_t = (\Delta_t^{\min}, \Delta_t^{\max})$. The model assumes $I=2$ or $I=3$ hidden layers and outputs our peak load prediction targets: $\Delta_t^{\min}$ and $\Delta_t^{\max}$. The model is regularized using input feature selection, a dropout layer, and $L_1$ regularization of the hidden layers and their weights. These methods are subject to tuning. We tune them along with other hyperparameters, such as the number of hidden layers, their activation functions, number of neurons, and the learning rate. We summarize all the hyperparameters to be tuned in the following list.
\begin{itemize}
	\item Input feature selection as shown in Tab.~\ref{tab_inputs} (21 hyperparameters).
	\item Number of hidden layers -- either 2 or 3 (1 hyperparameter).
	\item Dropout layer -- whether to use it after the input layer and if yes, at what rate. The rate is drawn uniformly from $(0,1)$ interval (up to 2~hyperparameters).
	\item Activation functions in the hidden layers. The possible functions are: elu, relu, sigmoid, softmax, softplus, and tanh (1~hyperparameter per layer).
	\item Number of neurons in the hidden layer drawn on an exp-scale from $[4,128]$ interval (1~hyperparameter per layer).
	\item $L_1$ regularization -- whether to use it on the hidden layers and their weights and if yes, at what rate. The rate is drawn on an exp-scale from $\left(10^{-5}, 10\right)$ interval (up to 4~hyperparameters per layer).
	\item Learning rate for the Adam optimization algorithm drawn on an exp-scale from $\left(10^{-5}, 10^{-1}\right)$ interval (1~hyperparameter).
\end{itemize}

This results in up to 43 hyperparameters that are subject to tuning. However, the number of possible combinations is far too high to check every single value for each parameter. Thus, we use the Optuna~\cite{akiba2019optuna} tool for the tuning exercise. {It utilizes a Bayesian optimization algorithm to find the optimal parameters.} The number of tuning iterations is set to 1000. Each iteration consists of a small, 3-steps expanding window study. In the first step, we consider in-sample data excluding the last three months, and we fit the model to that. Then, using the model, we predict the values for the following month. In the second step, we roll our window by one month, this time including the one that we predicted in the previous step, and repeat the fitting and predicting exercise. The third step is analogous to the second one. This way, we obtain the predictions of the recent three months of in-sample data with the given set of hyperparameters. To mimic the prediction study, we could tune the hyperparameters only using the most recent month. However, we use three of them for the sake of robustness, i.e., \ we want to avoid an overfitted deep neural network. After the whole iteration, the MSE of the predictions is calculated, and the Optuna sampler seeks the hyperparameter set that minimizes this value. The tuning procedure is repeated every month, i.e., before each modeling for the following month, resulting in 12 hyperparameter tunings.

The model is constructed and estimated with the Tensorflow~\cite{tensorflow2015-whitepaper} and Keras~\cite{chollet2015keras} framework. In addition to the tuned parameters, we arbitrarily set the validation set size to 25\% of the input data, the maximum number of epochs to 1500, and the batch size to $7\cdot48 = 336$. Moreover, we use the Adam optimizing algorithm, MSE as the loss function, and the early stopping callback with a patience of 50 epochs. Naturally, the input data is standardized prior to the modeling.

Due to the stochastic nature of the DNNs, we may face instability between multiple estimations of a single DNN. Therefore, the predictions are obtained each time using an ensemble of predictions provided by ten runs of each of the five best hyperparameter sets. {Ensembling is sometimes also called forecast averaging and has been applied to electricity price \cite{lago2021forecasting} and load forecasting \cite{takeda2016using} before. Thus, a single prediction represents an ensemble (here the mean) of 50 independent ones.}

\section{Empirical study and results}\label{results}

\subsection{GAM parameter significance}

\begin{table}
	\begin{center}
		\begin{tabular}{lrrlrr}
			\toprule
			\multicolumn{3}{c}{Minimum}                    & \multicolumn{3}{c}{Maximum}                                                                       \\
			\cmidrule(lr){1-3} \cmidrule(lr){4-6}
			Single Terms                                   & EDF                         & F    & Single Terms                                   & EDF  & F    \\
			\midrule
			$L_t$                                          & 9.5                         & 8.9  & $L_t$                                          & 10.8 & 17.6 \\
			$\widetilde{L}_{t-1}''$                        & 8.8                         & 4.4  & $\widetilde{L}_{t-2}''$                        & 6.7  & 4.9  \\
			$\widetilde{L}_{t+1}''$                        & 8.4                         & 4.8  & $L_{t-1}$                                      & 6.3  & 6.9  \\
			$\widetilde{L}_{t}''$                          & 8.2                         & 4.6  & $L_{t+1}$                                      & 5.4  & 5.0  \\
			Humid                                          & 6.1                         & 3.7  & $D_t$                                          & 4.8  & 3.1  \\
			WindE                                          & 5.1                         & 8.9  & Temp                                           & 4.1  & 11.5 \\
			$A_t$                                          & 4.7                         & 4.9  & Solar                                          & 4.0  & 5.5  \\
			WindN                                          & 4.2                         & 8.5  & WindE                                          & 3.8  & 5.8  \\
			Temp                                           & 3.5                         & 3.6  & $\widetilde{L}_{t+4}''$                        & 3.2  & 2.1  \\
			$L_{t-1}$                                      & 3.3                         & 1.3  & $\widetilde{L}_{t+1}''$                        & 3.1  & 2.2  \\
			\midrule
			Interactions                                   & EDF                         & F    & Interactions                                   & EDF  & F    \\
			\midrule
			$L_t$, $A_t$                                   & 26.7                        & 7.8  & $L_t$, $A_t$                                   & 37.8 & 28.1 \\
			$L_t$, Solar                                   & 20.7                        & 11.2 & $L_t$, $D_t$                                   & 25.1 & 20.6 \\
			$L_t$, $D_t$                                   & 18.3                        & 8.2  & Solar, $A_t$                                   & 15.4 & 6.4  \\
			$L_t$, $L_{t+1}$                               & 17.1                        & 4.0  & $\widetilde{L}_{t}''$, $\widetilde{L}_{t-1}''$ & 14.9 & 6.5  \\
			$\widetilde{L}_{t}''$, $\widetilde{L}_{t-1}''$ & 13.5                        & 2.5  & Solar, $D_t$                                   & 11.0 & 8.5  \\
			$L_t$, Temp                                    & 13.3                        & 3.1  & $\widetilde{L}_{t}''$, $\widetilde{L}_{t+1}''$ & 11.0 & 2.4  \\
			$L_t$, $L_{t-1}$                               & 13.0                        & 4.9  & $\widetilde{L}_{t}''$, $\widetilde{L}_{t+2}''$ & 10.7 & 2.8  \\
			$\widetilde{L}_{t}''$, $\widetilde{L}_{t+1}''$ & 12.1                        & 2.3  & $L_t$, Solar                                   & 10.0 & 2.9  \\
			$L_t$, $W_t$                                   & 11.7                        & 3.0  & $\widetilde{L}_{t}''$, Temp                    & 9.5  & 3.4  \\
			$\widetilde{L}_{t}''$, $\widetilde{L}_{t+2}''$ & 11.4                        & 2.9  & $L_t$, WindE                                   & 9.3  & 2.5  \\
			$L_t$, Humid                                   & 10.1                        & 2.6  & $L_t$, $\widetilde{L}_{t-4}''$                 & 9.1  & 2.1  \\
			$\widetilde{L}_{t}''$, $\widetilde{L}_{t-3}''$ & 9.9                         & 3.1  & $\widetilde{L}_{t}''$, $\widetilde{L}_{t-3}''$ & 9.0  & 2.2  \\
			$L_t$, WindN                                   & 9.9                         & 4.4  & Solar, WindE                                   & 9.0  & 2.5  \\
			$\widetilde{L}_{t}''$, Humid                   & 9.8                         & 1.8  & $L_t$, $W_t$                                   & 8.9  & 1.9  \\
			$\widetilde{L}_{t}''$, $\widetilde{L}_{t-2}''$ & 9.3                         & 2.1  & $L_t$, $\widetilde{L}_{t-3}''$                 & 8.7  & 1.5  \\
			$L_{t+1}$, Solar                               & 8.3                         & 1.3  & $L_t$, $L_{t+1}$                               & 8.4  & 1.3  \\
			Solar, $D_t$                                   & 8.2                         & 2.0  & $L_t$, $\widetilde{L}_{t+3}''$                 & 7.4  & 1.3  \\
			$L_t$, WindE                                   & 7.6                         & 2.5  & $L_t$, $\widetilde{L}_{t+1}''$                 & 6.8  & 1.5  \\
			Solar, WindE                                   & 6.9                         & 1.6  & $L_t$, $\widetilde{L}_{t-1}''$                 & 6.5  & 1.4  \\
			$\widetilde{L}_{t}''$, WindE                   & 6.9                         & 1.5  & $L_t$, $\widetilde{L}_{t+4}''$                 & 6.1  & 1.0  \\
			$L_t$, $\widetilde{L}_{t+1}''$                 & 6.2                         & 4.2  &                                                &      &      \\
			\bottomrule
		\end{tabular}
		\caption{ Effective degree of freedom and F-statistic of selected model terms of model \textbf{GAM.red}.}\label{edf_gam}
	\end{center}
\end{table}

Tab.~\ref{edf_gam} shows parameter significance estimates of the reduced GAM model \textbf{GAM.red}~\eqref{eq_small_GAM}. We only report single terms and interactions with cumulative effective degrees of freedom (EDF) larger than 3 for the single terms and larger than 6 for the interactions to keep the table concise. The table also reports the F-statistic for an F-test that evaluates the joint significance with respect to no effect of the model term. We do not report the p-value here since all variables included in Tab.~\ref{edf_gam} are very close to 0. Large EDF values and F-statistics indicate the high importance of the corresponding model term.

We observe that $L_t$ is by far the most important input as a single term, but also in the interactions. This holds for both minimum and maximum models. Besides, we also see that various non-linear effects, represented by the tensor interactions, receive high EDF values. Overall, minimum and maximum models differ in various respects. However, the (lagged and modified) second derivatives of the load $\widetilde{L}_{t+k}$ receive high EDF values in both models. A similar conclusion can be made for the weather variables and interactions with the weather. Solar is clearly the most important among the weather variables, but temperature, wind, and humidity also contribute. Those weather effects are known to have some relevance in load forecasting literature as well~\cite{xie2016relative, haben2019short, muzaffar2019short}.

\subsection{DNN Hyperparameter tuning results}

\begin{figure}[t!]
	\centering
	\resizebox{\textwidth}{!}{
		\input{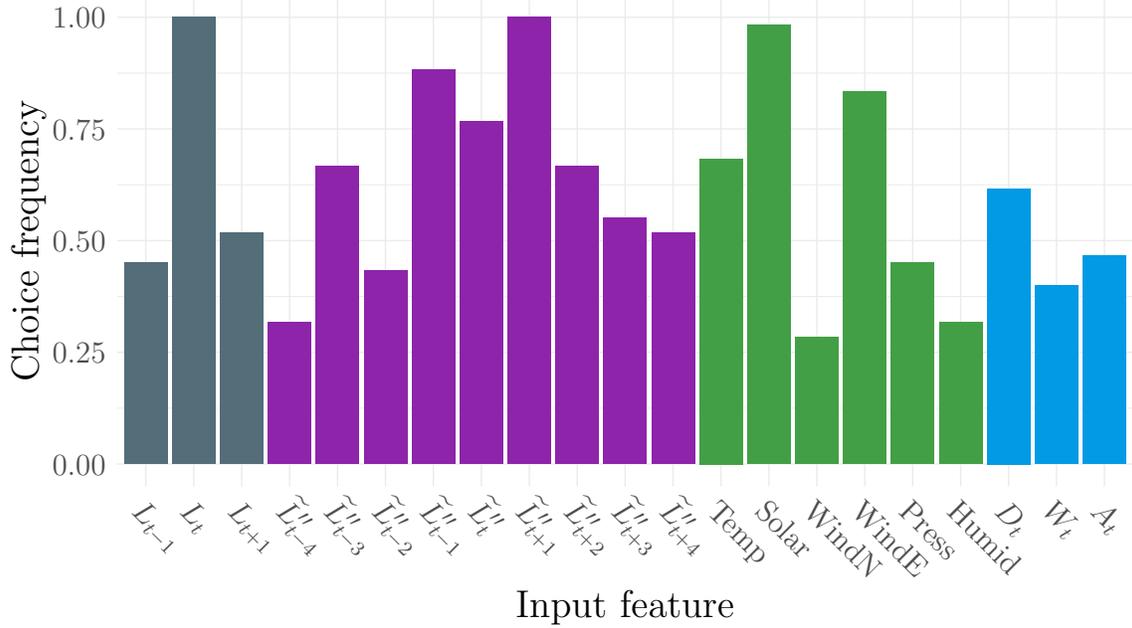}
	}
	\caption{Input choice frequency in the best 60 DNN hyperparameter sets.}\label{fig_DNN_input_frequency}
\end{figure}

We conduct the DNN hyperparameter tuning 12 times and each time we choose 5 best sets what results in 60 sets in total. Fig.~\ref{fig_DNN_input_frequency} presents the choice frequency of input features in the obtained sets.

\begin{figure}[t!]
	\centering
	\resizebox{\textwidth}{!}{
		\input{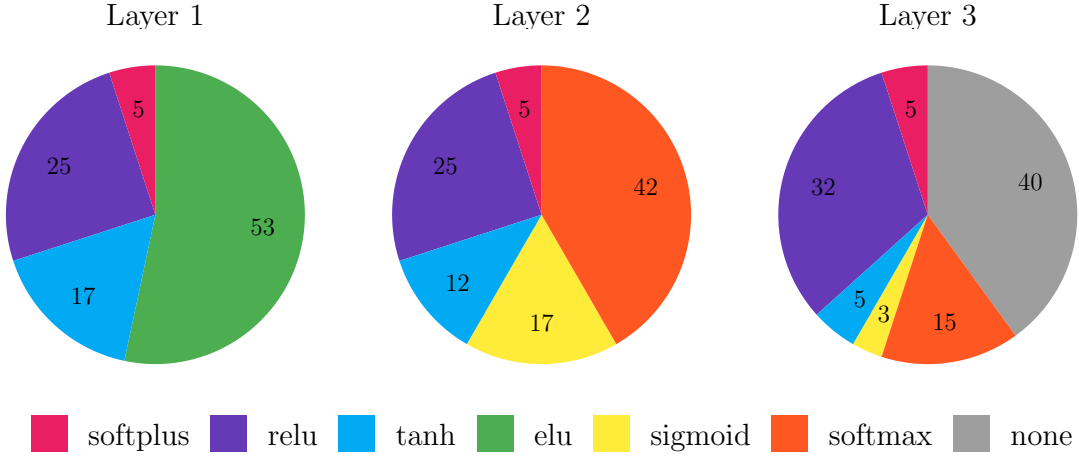}
	}
	\caption{{Share of chosen activation functions in the best 60 DNN hyperparameter sets in \%.}}\label{fig_DNN_activation_functions}
\end{figure}

\begin{figure}[t!]
	\centering
	\resizebox{\textwidth}{!}{
		\input{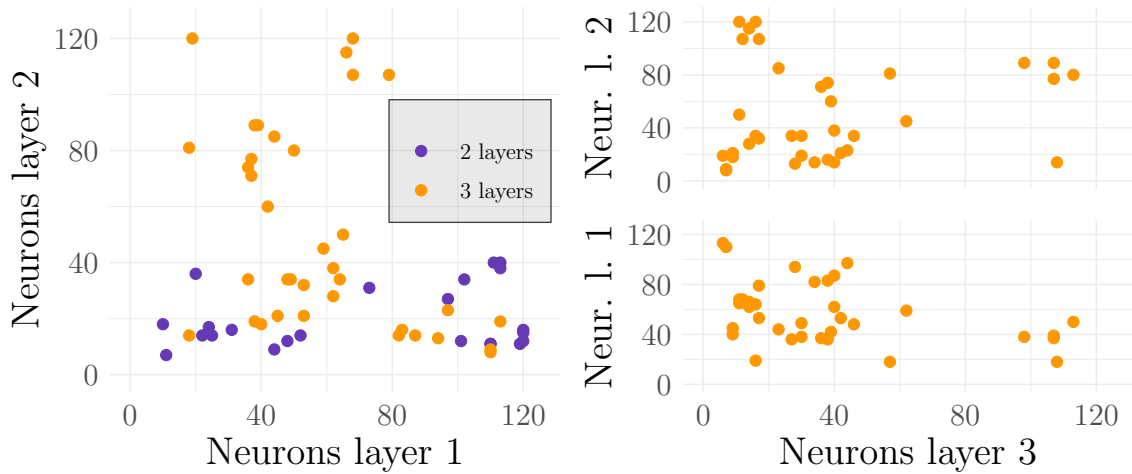}
	}
	\caption{{Neurons per layer in the best 60 DNN hyperparameter sets.}}\label{fig_neurons_per_layer}
\end{figure}

\begin{figure}[t!]
	\centering
	\resizebox{\textwidth}{!}{
		\input{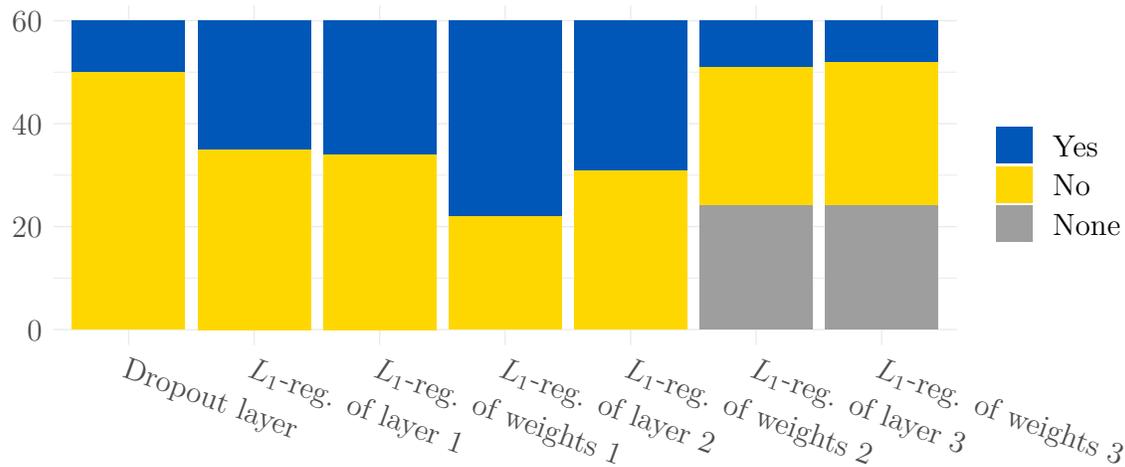}
	}
	\caption{{Regularization frequency in the best 60 DNN hyperparameter sets.}}\label{fig_regularization_dnn}
\end{figure}

The most important inputs are $L_t$, $\widetilde{L}_{t+1}''$ and Solar, which is similar to the parameter importance in the \textbf{GAM.red} model, see Tab.~\ref{edf_gam}. The least important variables are $\widetilde{L}_{t-4}''$ and WindN. The DNN uses three hidden layers in 60\% of cases and two in 40\% of cases. On average, the first hidden layer consists of 66 neurons, the second and the third (if present) of 38. {Fig.~\ref{fig_neurons_per_layer} presents the dependency between the numbers on different layers. Interestingly, the 2-layer networks tend to be smaller in the number of neurons than the 3-layer ones.}

	{Fig~\ref{fig_DNN_activation_functions} presents how often particular activation functions were chosen in each of the layers. We see that the elu function was chosen in half of the first layers, but in none of the remaining layers. The relu function is equally distributed across the first two layers and dominates the third layer. The most often chosen activation function in layer 2 is the softmax. The regularization frequency through dropout and $L_1$ regularization is reported in Fig.~\ref{fig_regularization_dnn}. We observe that, in most cases, the most optimal networks neglected these ways of regularization. Only the $L_1$ regularization of layer two was used in more than half of the tuned networks.}

\subsection{Study design}\label{subsec_study}
We evaluate the proposed models in an empirical evaluation study. The competitive models under consideration are the full GAM model \textbf{GAM.full}~\eqref{eq_big_GAM}, the reduced GAM model \textbf{GAM.red}~\eqref{eq_small_GAM}, and the deep neural network \textbf{DNN}~\eqref{eq_MLP}. Combining sophisticated models often improves predictive accuracy, thus we also consider the naive combination of~\textbf{GAM.full}, \textbf{GAM.red} and \textbf{DNN}. The naive combination uses uniform weights for the three models.

Additionally, we consider the simple GAM model \textbf{GAM.simple}~\eqref{eq_simple_GAM}, the full GAM model~\eqref{eq_big_GAM} without weather inputs \textbf{GAM.noWeather}, and the naive benchmark \textbf{Naive} defined by $\widetilde{L}_t^{\max} = L_t$ and $\widetilde{L}_t^{\min} = L_t$ for comparison purposes.

As mentioned, we evaluate the predictive accuracy of the models in a rolling window study, measuring the performance in the 12 months from 10/2020 to 09/2021. We evaluate the predictions by calculating the root mean squared error (RMSE) of the predictions. Here, we distinguish a joint RMSE and an RMSE for the minimum and maximum:
\begin{align}
	\text{RMSE}     & = \sqrt{ \sum_{t\in \mathbb T} (\widetilde{L}_t^{\min} - L_t^{\min})^2 + (\widetilde{L}_t^{\max} - L_t^{\max})^2 } \\
	\text{RMSE}_{m} & = \sqrt{ \sum_{t\in \mathbb T} (\widetilde{L}_t^{m} - L_t^{m})^2  }                                                
\end{align}
where $m\in\{\min, \max \}$ and $\mathbb T$ represents the index set of time points of the considered month.
The naive benchmark model was also used in the competition skill score which corresponds to the relative improvement in the RMSE:
\begin{align}
	\text{Score} & = \text{RMSE}(\text{Model}) / \text{RMSE}(\text{\textbf{naive}}).
\end{align}

\subsection{Evaluation results}

\begin{table*}
	\includegraphics[trim={4cm 3cm 4cm 0cm},clip, width=\textwidth]{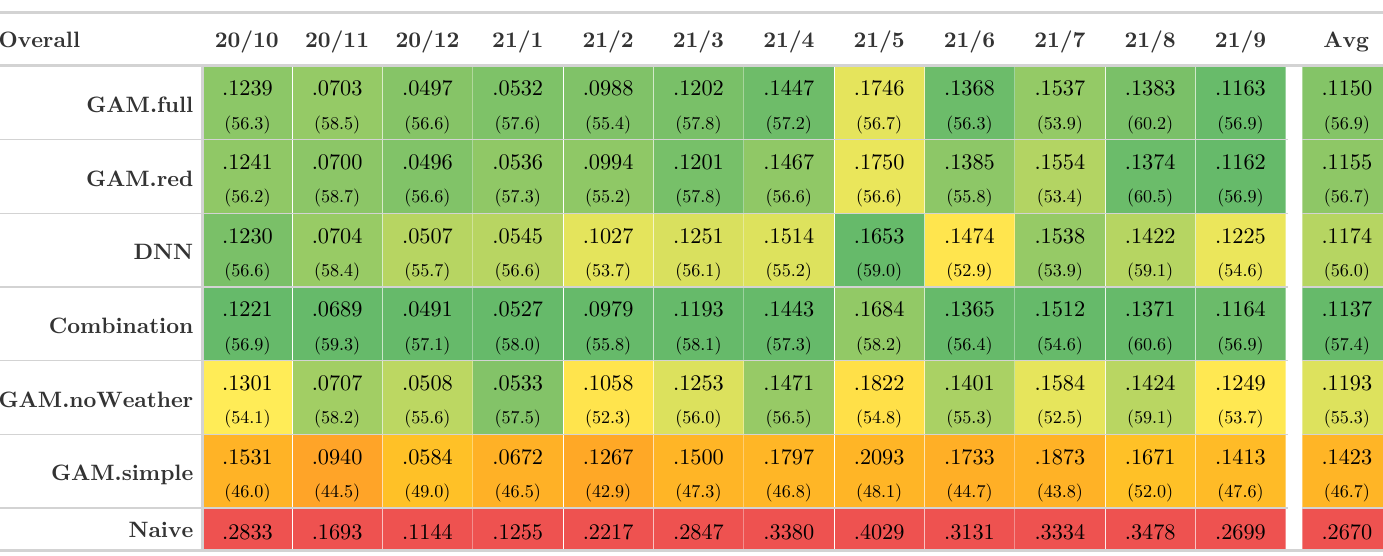}
	\caption{{RMSE values for peak load prediction in MW in the evaluation study with corresponding averages (last column). Relative improvements over the \textbf{Naive} model in \% are in brackets below.}}\label{tab_results_main}
\end{table*}

Tab.~\ref{tab_results_main} shows the RMSE and corresponding score results of the considered models. Overall, we see that all sophisticated models show high predictive accuracy. The \textbf{GAM.full} and \textbf{GAM.red} have almost the same predictive accuracy and improve the RMSE of \textbf{naive} by about 57\%. The \textbf{DNN} model performs only slightly worse with an accuracy improvement of 56\%. Still, we see a few months where \textbf{DNN} outperforms the GAM models, namely October 2020 and May 2021. Hence, the \textbf{Combination} has a robust predictive accuracy and exhibits another small improvement over the individual models. This improvement is about 57.5\% compared to \textbf{naive}. We want to mention that in the WPD challenge, the test month was September 2021. Here, the GAM  models performed particularly well with a skill score of 43.1\% (56.9\% Improvement compared to \textbf{naive}) - the same score as the more robust \textbf{Combination} model. In the competition, we used a slightly modified combination model (e.g., the GAM also included manually selected 3-way interactions) with a final skill score of 42.6\% (57.4\% Improvement compared to \textbf{naive}). The second and third-placed teams (\textit{WOJJ} and \textit{code\_green}) received a score of 43.6\% and 43.7\%. Thus, the proposed \textbf{Combination} model has an improvement of more than 1 percentage point in terms of RMSE compared to the other teams.

From the \textbf{GAM.noWeather} model in Tab.~\ref{tab_results_main} we can deduce that the weather variables improve the RMSE skill score by 1.6 percentage points on average. This sounds minor, but it is substantial, considering that the difference in the skill score between \textbf{GAM.simple} and \textbf{GAM.full} is only about 10 percentage points.

Tab.~\ref{tab_results_min_max} presents the $\text{RMSE}_{\min}$ and $\text{RMSE}_{\max}$ results. We observe that the performance between the GAM models and the DNN varies with the prediction target. For the minimum peak load models, the DNN performs clearly worse than the GAM models, even worse than the \textbf{GAM.noWeather} model. Thus, the predictive accuracy of the  \textbf{Combination} model improves only marginally compared to the well-performing GAM models \textbf{GAM.full} and \textbf{GAM.red}. In contrast, the neural network model performs similarly to the two GAM models for the maximum peak load. Thus, we see a substantial increase in predictive accuracy due to aggregation in the \textbf{Combination} model.

\begin{table*}
	\includegraphics[trim={4cm 4.4cm 4cm 0cm},clip, width=\textwidth]{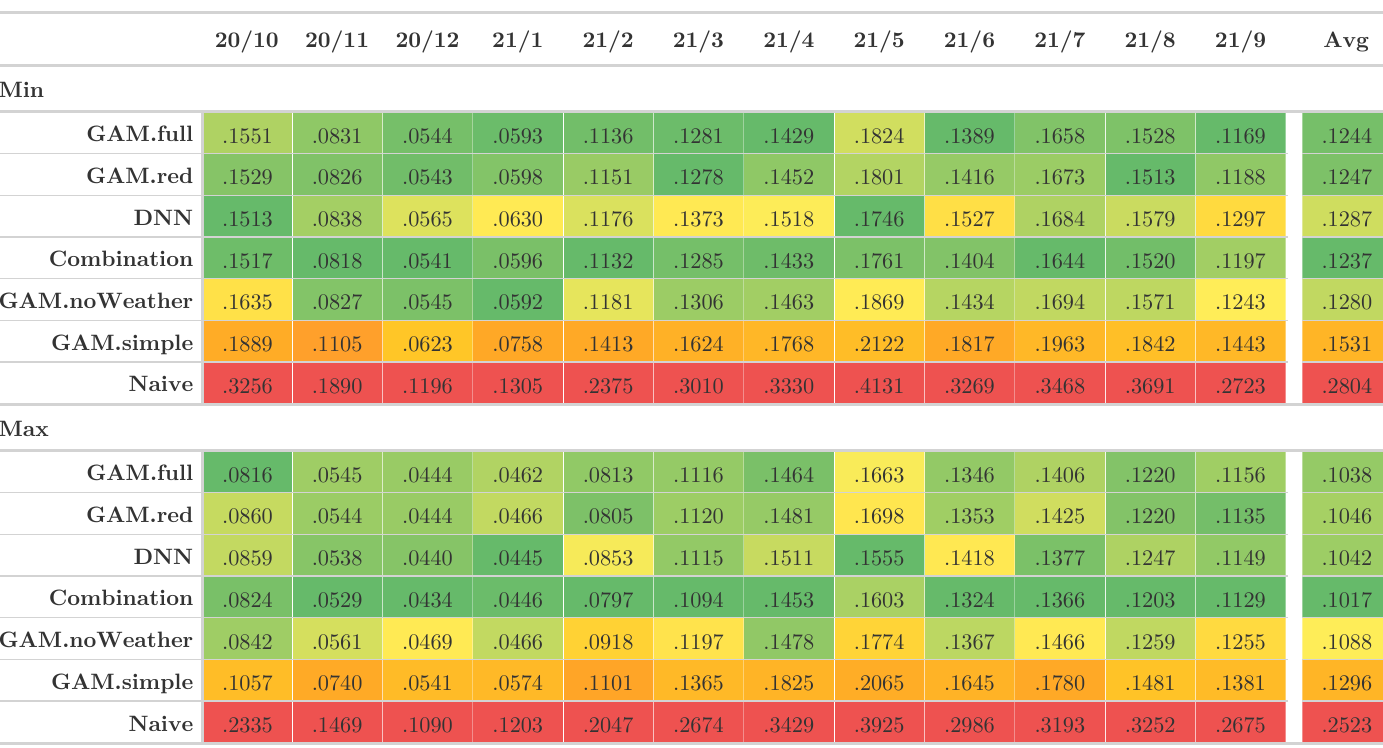}
	\caption{$\text{RMSE}_{\min}$ and $\text{RMSE}_{\max}$ values for the minimum and maximum peak load in MW for considered prediction models in the evaluation study with corresponding average (last column).}\label{tab_results_min_max}
\end{table*}


\section{Conclusion}\label{conclusion}

This paper covers estimating high-resolution electricity peak demand given lower resolution data. For this purpose, we developed models using two distinct model classes, namely \textbf{GAM} and \textbf{DNN}. Combining these models received the best score in the WPD data competition. The month to predict in the competition was September 2021. However, to evaluate the robustness of our proposed methodology, we report model performance for eleven additional months. The model components and their importance were discussed in detail.
Given the results, we conclude that the most important ones are the load $L_t$, its modified second derivative $\widetilde{L}_t''$, and Solar. This applies to both \textbf{GAM} and \textbf{DNN} models. The proposed combination of both model classes improves performance and delivers more robust predictions. The method was proven in a robust 12-months evaluation study. Overall, this combination yields an RMSE that is 1.2\% lower compared to the best performing individual model \textbf{GAM.full} and 57.4\% lower compared to the \textbf{Naive} benchmark. {Including weather information was clearly beneficial. Comparing \textbf{GAM.full} to \textbf{GAM.noWeather} shows that considering weather variables improves the RMSE by 1.6 percentage points on average. The proposed methodology can easily be applied to different substations. An interesting question is how the model performance changes when the data is from a different region. If the results generalize well, network operators would not need to pay for collecting high-resolution data in the first place. The high-resolution features of interest could be extracted from low-resolution data using the proposed methods. This is an interesting topic for future work.}

\bibliographystyle{anc/elsarticle-num.bst}
\bibliography{references}

\begin{thebibliography}{10}
\expandafter\ifx\csname url\endcsname\relax
  \def\url#1{\texttt{#1}}\fi
\expandafter\ifx\csname urlprefix\endcsname\relax\def\urlprefix{URL }\fi
\expandafter\ifx\csname href\endcsname\relax
  \def\href#1#2{#2} \def\path#1{#1}\fi

\bibitem{zheng2013smart}
J.~Zheng, D.~W. Gao, L.~Lin, Smart meters in smart grid: An overview, in: 2013
  IEEE Green Technologies Conference (GreenTech), IEEE, 2013, pp. 57--64.

\bibitem{suanduleac2021high}
M.~S{\u{a}}nduleac, I.~Ciornei, L.~Toma, R.~Pl{\u{a}}mnescu, A.-M. Dumitrescu,
  M.~M. Albu, High reporting rate smart metering data for enhanced grid
  monitoring and services for energy communities, IEEE Transactions on
  Industrial Informatics 18~(6) (2021) 4039--4048.

\bibitem{mughees2021deep}
N.~Mughees, S.~A. Mohsin, A.~Mughees, A.~Mughees, Deep sequence to sequence
  bi-lstm neural networks for day-ahead peak load forecasting, Expert Systems
  with Applications 175 (2021) 114844.

\bibitem{lee2022national}
J.~Lee, Y.~Cho, {National-scale electricity peak load forecasting: Traditional,
  machine learning, or hybrid model?}, Energy 239 (2022) 122366.

\bibitem{uddin2018review}
M.~Uddin, M.~F. Romlie, M.~F. Abdullah, S.~Abd~Halim, T.~C. Kwang, et~al., A
  review on peak load shaving strategies, Renewable and Sustainable Energy
  Reviews 82 (2018) 3323--3332.

\bibitem{lissa2021deep}
P.~Lissa, C.~Deane, M.~Schukat, F.~Seri, M.~Keane, E.~Barrett, Deep
  reinforcement learning for home energy management system control, Energy and
  AI 3 (2021) 100043.

\bibitem{sun2018probabilistic}
M.~Sun, Y.~Wang, G.~Strbac, C.~Kang, Probabilistic peak load estimation in
  smart cities using smart meter data, IEEE Transactions on Industrial
  Electronics 66~(2) (2018) 1608--1618.

\bibitem{chou2018forecasting}
J.-S. Chou, D.-S. Tran, {Forecasting energy consumption time series using
  machine learning techniques based on usage patterns of residential
  householders}, Energy 165 (2018) 709--726.

\bibitem{chen2017short}
Y.~Chen, P.~Xu, Y.~Chu, W.~Li, Y.~Wu, L.~Ni, Y.~Bao, K.~Wang, {Short-term
  electrical load forecasting using the Support Vector Regression (SVR) model
  to calculate the demand response baseline for office buildings}, Applied
  Energy 195 (2017) 659--670.

\bibitem{haben2019short}
S.~Haben, G.~Giasemidis, F.~Ziel, S.~Arora, Short term load forecasting and the
  effect of temperature at the low voltage level, International Journal of
  Forecasting 35~(4) (2019) 1469--1484.

\bibitem{guo2018deep}
Z.~Guo, K.~Zhou, X.~Zhang, S.~Yang, A deep learning model for short-term power
  load and probability density forecasting, Energy 160 (2018) 1186--1200.

\bibitem{xie2016relative}
J.~Xie, Y.~Chen, T.~Hong, T.~D. Laing, Relative humidity for load forecasting
  models, IEEE Transactions on Smart Grid 9~(1) (2016) 191--198.

\bibitem{dehalwar2016electricity}
V.~Dehalwar, A.~Kalam, M.~L. Kolhe, A.~Zayegh, Electricity load forecasting for
  urban area using weather forecast information, in: 2016 IEEE International
  Conference on Power and Renewable Energy (ICPRE), IEEE, 2016, pp. 355--359.

\bibitem{cai2019day}
M.~Cai, M.~Pipattanasomporn, S.~Rahman, {Day-ahead building-level load
  forecasts using deep learning vs. traditional time-series techniques},
  Applied energy 236 (2019) 1078--1088.

\bibitem{muzaffar2019short}
S.~Muzaffar, A.~Afshari, {Short-term load forecasts using LSTM networks},
  Energy Procedia 158 (2019) 2922--2927.

\bibitem{acarouglu2021comprehensive}
H.~Acaro{\u{g}}lu, F.~P. Garc{\'\i}a~M{\'a}rquez, Comprehensive review on
  electricity market price and load forecasting based on wind energy, Energies
  14~(22) (2021) 7473.

\bibitem{hong2015weather}
T.~Hong, P.~Wang, L.~White, Weather station selection for electric load
  forecasting, International Journal of Forecasting 31~(2) (2015) 286--295.

\bibitem{vu2017short}
D.~H. Vu, K.~M. Muttaqi, A.~P. Agalgaonkar, A.~Bouzerdoum, Short-term
  electricity demand forecasting using autoregressive based time varying model
  incorporating representative data adjustment, Applied Energy 205 (2017)
  790--801.

\bibitem{aguilar2021short}
E.~Aguilar~Madrid, N.~Antonio, Short-term electricity load forecasting with
  machine learning, Information 12~(2) (2021) 50.

\bibitem{zhang2017short}
X.~Zhang, J.~Wang, K.~Zhang, {Short-term electric load forecasting based on
  singular spectrum analysis and support vector machine optimized by Cuckoo
  search algorithm}, Electric Power Systems Research 146 (2017) 270--285.

\bibitem{sheng2020short}
F.~Sheng, L.~Jia, {Short-term load forecasting based on SARIMAX-LSTM}, in: 2020
  5th International Conference on Power and Renewable Energy (ICPRE), IEEE,
  2020, pp. 90--94.

\bibitem{fan2021forecasting}
G.-F. Fan, M.~Yu, S.-Q. Dong, Y.-H. Yeh, W.-C. Hong, Forecasting short-term
  electricity load using hybrid support vector regression with grey catastrophe
  and random forest modeling, Utilities Policy 73 (2021) 101294.

\bibitem{pierrot2011short}
A.~Pierrot, Y.~Goude, {Short-term electricity load forecasting with generalized
  additive models}, Proceedings of ISAP power 2011.

\bibitem{goude2013local}
Y.~Goude, R.~Nedellec, N.~Kong, {Local short and middle term electricity load
  forecasting with semi-parametric additive models}, IEEE transactions on smart
  grid 5~(1) (2013) 440--446.

\bibitem{ziel2022smoothed}
F.~Ziel, Smoothed bernstein online aggregation for short-term load forecasting
  in ieee dataport competition on day-ahead electricity demand forecasting:
  Post-covid paradigm, IEEE Open Access Journal of Power and Energy.

\bibitem{amato2021forecasting}
U.~Amato, A.~Antoniadis, I.~De~Feis, Y.~Goude, A.~Lagache, {Forecasting high
  resolution electricity demand data with additive models including smooth and
  jagged components}, International Journal of Forecasting 37~(1) (2021)
  171--185.

\bibitem{tasre2011daily}
M.~B. Tasre, P.~P. Bedekar, V.~N. Ghate, {Daily peak load forecasting using
  ANN}, in: 2011 Nirma University International Conference on Engineering,
  IEEE, 2011, pp. 1--6.

\bibitem{pallonetto2022forecast}
F.~Pallonetto, C.~Jin, E.~Mangina, Forecast electricity demand in commercial
  building with machine learning models to enable demand response programs,
  Energy and AI 7 (2022) 100121.

\bibitem{hosein2017load}
S.~Hosein, P.~Hosein, {Load forecasting using deep neural networks}, in: 2017
  IEEE Power \& Energy Society Innovative Smart Grid Technologies Conference
  (ISGT), IEEE, 2017, pp. 1--5.

\bibitem{amarasinghe2017deep}
K.~Amarasinghe, D.~L. Marino, M.~Manic, Deep neural networks for energy load
  forecasting, in: 2017 IEEE 26th International Symposium on Industrial
  Electronics (ISIE), IEEE, 2017, pp. 1483--1488.

\bibitem{khwaja2020joint}
A.~S. Khwaja, A.~Anpalagan, M.~Naeem, B.~Venkatesh, {Joint bagged-boosted
  artificial neural networks: Using ensemble machine learning to improve
  short-term electricity load forecasting}, Electric Power Systems Research 179
  (2020) 106080.

\bibitem{walser2021typical}
T.~Walser, A.~Sauer, Typical load profile-supported convolutional neural
  network for short-term load forecasting in the industrial sector, Energy and
  AI 5 (2021) 100104.

\bibitem{memarzadeh2021short}
G.~Memarzadeh, F.~Keynia, Short-term electricity load and price forecasting by
  a new optimal lstm-nn based prediction algorithm, Electric Power Systems
  Research 192 (2021) 106995.

\bibitem{shaqour2022electrical}
A.~Shaqour, T.~Ono, A.~Hagishima, H.~Farzaneh, Electrical demand aggregation
  effects on the performance of deep learning-based short-term load forecasting
  of a residential building, Energy and AI 8 (2022) 100141.

\bibitem{bashir2022short}
T.~Bashir, C.~Haoyong, M.~F. Tahir, Z.~Liqiang, Short term electricity load
  forecasting using hybrid prophet-lstm model optimized by bpnn, Energy Reports
  8 (2022) 1678--1686.

\bibitem{khan2022efficient}
Z.~A. Khan, A.~Ullah, I.~U. Haq, M.~Hamdy, G.~M. Maurod, K.~Muhammad, M.~Hijji,
  S.~W. Baik, Efficient short-term electricity load forecasting for effective
  energy management, Sustainable Energy Technologies and Assessments 53 (2022)
  102337.

\bibitem{gneiting2011making}
T.~Gneiting, {Making and evaluating point forecasts}, Journal of the American
  Statistical Association 106~(494) (2011) 746--762.

\bibitem{szekely2007measuring}
G.~J. Sz{\'e}kely, M.~L. Rizzo, N.~K. Bakirov, Measuring and testing dependence
  by correlation of distances, The annals of statistics 35~(6) (2007)
  2769--2794.

\bibitem{ziel2021m5}
F.~Ziel, M5 competition uncertainty: Overdispersion, distributional
  forecasting, gamlss, and beyond, International Journal of Forecasting.

\bibitem{marz2022distributional}
A.~M{\"a}rz, T.~Kneib, Distributional gradient boosting machines, arXiv
  preprint arXiv:2204.00778.

\bibitem{wood2017gam}
S.~Wood, {Generalized Additive Models: An Introduction with R}, 2nd Edition,
  Chapman and Hall/CRC, 2017.

\bibitem{akiba2019optuna}
T.~Akiba, S.~Sano, T.~Yanase, T.~Ohta, M.~Koyama, {Optuna: A next-generation
  hyperparameter optimization framework}, in: Proceedings of the 25th ACM
  SIGKDD international conference on knowledge discovery \& data mining, 2019,
  pp. 2623--2631.

\bibitem{tensorflow2015-whitepaper}
M.~Abadi, A.~Agarwal, P.~Barham, E.~Brevdo, Z.~Chen, C.~Citro, G.~S. Corrado,
  A.~Davis, J.~Dean, M.~Devin, S.~Ghemawat, I.~Goodfellow, A.~Harp, G.~Irving,
  M.~Isard, Y.~Jia, R.~Jozefowicz, L.~Kaiser, M.~Kudlur, J.~Levenberg,
  D.~Man\'{e}, R.~Monga, S.~Moore, D.~Murray, C.~Olah, M.~Schuster, J.~Shlens,
  B.~Steiner, I.~Sutskever, K.~Talwar, P.~Tucker, V.~Vanhoucke, V.~Vasudevan,
  F.~Vi\'{e}gas, O.~Vinyals, P.~Warden, M.~Wattenberg, M.~Wicke, Y.~Yu,
  X.~Zheng, \href{https://www.tensorflow.org/}{{TensorFlow}: {Large-Scale
  Machine Learning on Heterogeneous Systems}}, software available from
  tensorflow.org (2015).
\newline\urlprefix\url{https://www.tensorflow.org/}

\bibitem{chollet2015keras}
F.~Chollet, et~al., Keras, \url{https://keras.io} (2015).

\bibitem{lago2021forecasting}
J.~Lago, G.~Marcjasz, B.~De~Schutter, R.~Weron, Forecasting day-ahead
  electricity prices: A review of state-of-the-art algorithms, best practices
  and an open-access benchmark, Applied Energy 293 (2021) 116983.

\bibitem{takeda2016using}
H.~Takeda, Y.~Tamura, S.~Sato, Using the ensemble kalman filter for electricity
  load forecasting and analysis, Energy 104 (2016) 184--198.

\end{thebibliography}

\end{document}